\documentclass[runningheads]{llncs}
\usepackage{graphicx}
\usepackage[dvipsnames]{xcolor} 
\usepackage{algorithm}
\usepackage{algpseudocode}
\usepackage{amssymb}
\usepackage{amsmath}
\usepackage{xfrac}

\newif\ifanonymized

\begin{document}

%\anonymizedtrue 
\anonymizedfalse

\title{LumiPath -- Towards Real-time Physically-based Rendering on Embedded Devices}
\titlerunning{LumiPath}

\ifanonymized
\else
\author{Laura~Fink$^{1,2}$, Sing~Chun~Lee$^1$, Jie Ying Wu$^1$, Xingtong Liu$^1$, Tianyu Song$^1$, Yordanka Stoyanova$^1$, Marc~Stamminger$^2$, Nassir~Navab$^1$, and Mathias~Unberath$^1$}
\authorrunning{Fink et al.}

% index{Fink, Laura}
% index{Lee, Sing Chun}
% index{Wu, Jie Ying}
% index{Liu, Xingtong}
% index{Song, Tianyu}
% index{Stoyanova, Yordanka}
% index{Stamminger, Marc}
% index{Navab, Nassir}
% index{Unberath, Mathias}

\institute{
$^1$ Johns Hopkins University\\
$^2$ Friedrich-Alexander University Erlangen-Nuremberg
}

\fi
\maketitle              % typeset the header of the contribution
%

% Disclaimer paragraph described is work in progress

% Make the volume rendering more abstract - like getting from gaming to medicine e.g. 

\begin{abstract}
With the increasing computational power of today's workstations, real-time physi\-cally-based rendering is within reach, ra\-pid\-ly gaining attention across a variety of domains. %These range from gaming, where physically-based rendering enhances immersion and overall experience, to medicine, 
These have expeditiously applied to medicine, where it is a powerful tool for intuitive 3D data visualization.  Embedded devices such as optical see-through head-mounted displays~(OST HMDs) have been a trend for medical augmented reality.  However, leveraging the obvious benefits of physically-based rendering remains challenging on these devices because of limited computational power, memory usage, and power consumption. We navigate the compromise between device limitations and image quality to achieve reasonable rendering results by introducing a novel light field that can be sampled in real-time on embedded devices.
We demonstrate its applications in medicine and discuss limitations of the proposed method. An open-source version of this project is available at \texttt{\small \url{https://github.com/lorafib/LumiPath}} which provides full insight on implementation and exemplary demonstrational material.
\end{abstract}

\section{Introduction}
\noindent\textbf{Real-time Physically-based Rendering }
Conventional rasterization methods generate images by artificially shading objects but mostly limit considerations to direct illumination. In contrast, physically-based rendering aims to synthesize images by simulating light propagation. To this end, these methods consider how light quanta are emitted from light sources and interact with the environment before impinging on a camera's image plane.
As a direct consequence, physically-based rendering additionally provides indirect illumination effects which have a high impact on perceived realism. 
One such method, \textit{ray tracing}~\cite{Rademacher1997}, simulates light rays in a reverse order. 
Incoming radiance is integrated for each pixel by following rays that are emitted from the camera. These rays hit objects in the scene which they interact with, based on the physical simulation of illumination phenomena such as reflection, refraction, and shadowing. 
From these hit-points, again all incoming radiance is integrated and rays are repeatedly traced until they eventually reach a light source (or exit the scene). 

Accurately accounting for imaging physics can result in rendered images that are indiscernible from real ones. 
However, integrating incoming radiance for each pixel is computationally expensive and barely real-time.  
Hardware, like the \textit{Nvidia GeForce RTX}, made a big step towards real-time ray tracing by incorporating deep learning technology, drastically reducing the required computations. This is achieved by aggressively limiting light-scene interactions; resulting artifacts are masked with machine learning-based post-processing.

The increase in compute capabilities of graphics processing units (GPUs) and advances of rendering algorithms have fueled the recent interest in adopting real-time physically-based rendering in daily applications. 
Unfortunately, these advances do not translate well to applications on embedded devices. 
This is because 1) GPU hardware cannot easily be miniaturized and integrated, and 2) remote-computation and streaming is not necessarily desirable (particularly in the medical context). In the remainder of this manuscript, we describe methods that aim at bringing real-time physically-based rendering to embedded devices.

\vspace{3mm}
\noindent\textbf{Related Work } We limit our non-exhaustive review of related work to plenoptic functions (light fields), and physically-based rendering on embedded devices.

\vspace{3mm}
\noindent \emph{The Plenoptic Function }  
Light transport in a 3D static scene can be expressed as tracing the set of all possible rays; rays are defined by their origin $(x,y,z) \in \mathbb{R}^3$ and direction $(\theta, \phi) \in [0, \pi]\times[0, 2\pi]$, yielding five degrees of freedom~(DoFs). 
This description is referred to as \textit{light field} %~\cite{gershun1939light}
or \textit{plenoptic function}%~\cite{AdelsonMIT1991,michael1846ray}
. Hardware limitations led to precomputing a subset of the plenoptic function in a domain of interest rather than  
simulating light-scene interaction
on the fly. 
Image synthetization is then performed by sampling and interpolating the  precomputed results. Such approaches are referred to as \textit{image based rendering}~\cite{mcmillan1995plenoptic} and are capable of highly reducing the computations needed at runtime.

Among the most well-known representatives of such approaches is the  LumiGraph~\cite{GortlerSIGGRAPH1996}, which reduces the five~DoFs of the plenoptic function to four. 
The LumiGraph is based on the assumption that the medium surrounding an object of interest is transparent (radiance is constant along the ray), and therefore, the plenoptic function can be parameterized in terms of a bounding surface, namely a cube. By heavily constraining possible camera-object arrangements, this surface can be further reduced by only considering two opposite sides of the cube, i.\,e. two planes.
Two point sets $\textrm{P}_o$ and $\textrm{P}_d$ discretize the first and second plane, respectively. 
The set of precomputed rays can then be determined by connecting every point $p_o \in \textrm{P}_o$ with each point $p_d \in \textrm{P}_d$. This arrangement may lead to artifacts~\cite{CamahortEuroGraphics1998} due to a non-uniform sampling of the light field.

Camahort et al.~\cite{CamahortEuroGraphics1998} examined sampling on a sphere to provide more uniform light fields. 
They perform a binning approach based on a Bresenham-style discretization of the spherical surface which, in addition to not being perfectly uniform, has 
the major drawback of the runtime complexity or radiance information query being dependent on the number of bins. 

\vspace{3mm}
\noindent \emph{Physically-based Rendering on Embedded Devices } 
A patent~\cite{zhou2018method} % and a press release~\cite{siemensHealthi}, both 
out of Siemens Healthineers is one of the closest works we are aware of that aims to achieve physically-based rendering on embedded devices.
Their method is similar to ours in that it partly front-loads computations to accelerate image generation. However, the application still seems to depend on ray casting at runtime which is found to be a quite demanding task for today's embedded devices, including head-mounted displays, in its own right~\cite{hajek2018closing}.

%\MS{This one goes into a similar direction: \texttt{https://dl.acm.org/citation.cfm?doid=2342896.2343040}}.
%\Laura{Ouch.}

%Attenuation Field~\cite{RussakoffTMI2005}
%Attenuation Box~\cite{GhafurianSPIE2016}

\vspace{3mm}
\noindent\textbf{Contributions}
In summary, our contributions are:
\vspace{-1mm}
\begin{itemize}
    \item An algorithm for real-time physically-based rendering-like results on embedded devices based on uniformly sampled light fields, which, to the best of our knowledge, is the first algorithm to do so.
    \item A new 2D plenoptic function representation using two \textit{Spherical Fibonacci point sets}~\cite{MarquesCGForum2013}, which are sampled uniformly and with arbitrary sampling size providing flexibility in tweaking memory to any embedded device. 
    \item Fast neighborhood query of our domain using an extended version of the \textit{Keinert Inverse Fibonacci Mapping}~\cite{KeinertACMTG2015}. It has constant time complexity per pixel, does not require additional query structures and is decoupled from the light field's discretization granularity. The runtime only depends on the fixed number of queried neighbors needed during color interpolation.
    \item An effective machine learning-based post processing filter which is well designed for the execution on embedded devices that trend to incorporate inferencing acceleration and its evaluation.
\end{itemize}

\section{Method}
\label{sec:method}
\begin{figure}
    \centering
    \begin{minipage}[c]{0.9\linewidth}
    \includegraphics[width=\textwidth]{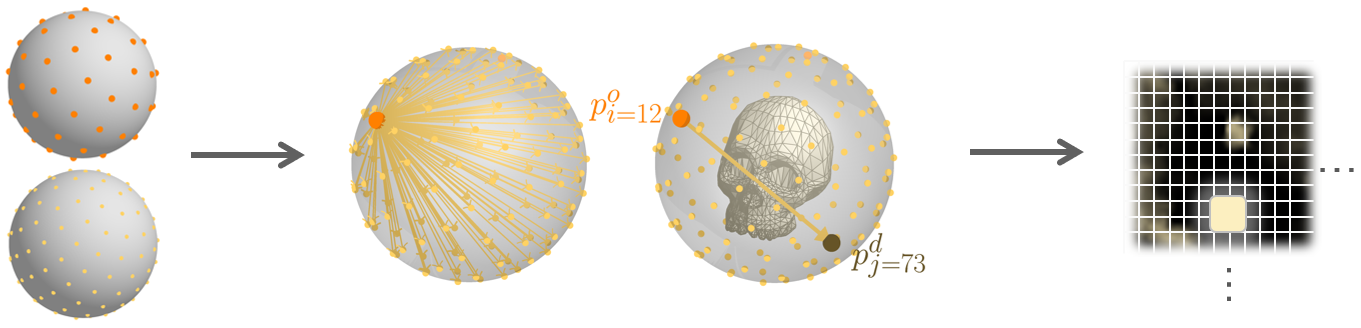}
    \end{minipage}
    \caption{$\hat{L}(i,j)$ and filling the texture. 
    The surface of the bounding sphere~\textbf{S} is discretized by the two point sets~$\textrm{P}^M_{o}$ and ~$\textrm{P}^N_{d}$. 
    Rays are traced from each $p^o_i$ to each $p^d_j$ resulting in re-parameterization and discretization of the plenoptic function, referred to as $\hat{L}(i,j)$.
    The value of $\hat{L}(i,j)$ is written to a 2D~texture at position~($i, j$).
    }
    \label{fig:sampling}
\end{figure}

\begin{figure}
    \centering
    \begin{minipage}[c]{\linewidth}
    \includegraphics[width=0.95\textwidth]{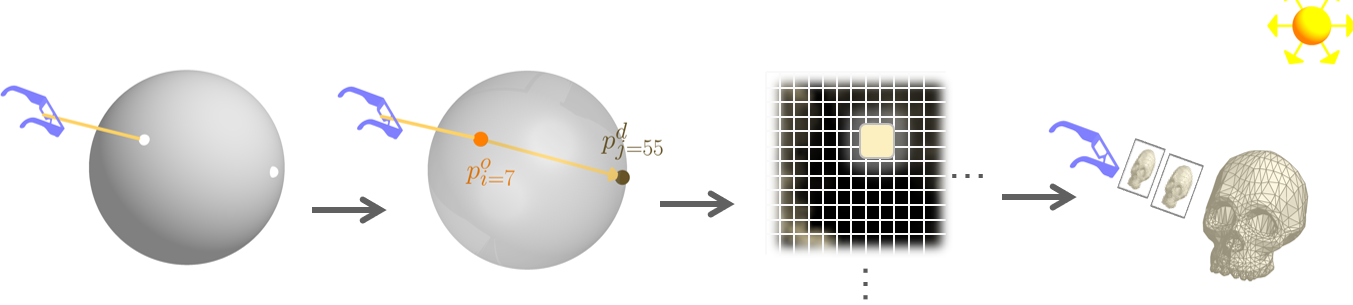}
    \end{minipage}
    \caption{Process of image synthetization by sampling $\hat{L}(i,j)$. A rasterization ray yields two hitpoints~$h_o$ and $h_d$~(white dots). In case of nearest sampling, $h_o$ and $h_d$ are mapped to their nearest neighbors $p^o_i$ and $p^d_j$. $(i,j)$ in the point sets~$\textrm{P}_o^M$ and $h_d$ from $\textrm{P}_d^N$ are used as coordinates to fetch a texel from the texture, and thus sampling $\hat{L}(i,j)$. 
    The mapping is performed for each pixel of the displayed image.}
    \label{fig:image_generation}
\end{figure}

In order to allow for physically-based rendering on embedded devices, our prototype consists of a two-step algorithm.
First, we compute all values of a reformulated plenoptic function and save the outcome as texture, which trades off hardware resources for rendering quality (see Fig.~\ref{fig:sampling}). 
Second, we transform the computationally expensive rendering task into a fast data query and interpolation task using this new representation (see Fig.~\ref{fig:image_generation}).
Additionally, we present a neural network that performs post-rendering correction in order to resolve artifacts and vastly enhance image quality.

\vspace{3mm}
\noindent\textbf{LumiPath-based Rendering } The parameterization of the plenoptic function~$L(x,y,z, \theta, \phi)$ implies that we consider our scene as static.
Further, we assume that the medium outside of a bounding sphere~\textbf{S} which encapsulates our domain of interest~(DOI) is totally transparent. Thus, radiance along a ray remains constant and consequently, the radiance emitted from the DOI is equal to the radiance at the intersection point of a ray with \textbf{S}.
We reparameterize and discretize the plenoptic function~$L$ according to the surface of \textbf{S} (with radius $R$ and origin $O$) and hence reduce the domain of $L$ to our DOI. 

For uniform discretization of the surface of~\textbf{S}, we use \textit{Spherical Fibonacci~(SF)} \textit{point sets}~$\textrm{P}^n_{S\!F}$~\cite{MarquesCGForum2013}.
A point $p_i$ of a SF point set $\textrm{P}^n_{S\!F}$ is given by 
\begin{equation*}
    p_i = C(\phi_i, \cos^{-1}(z_i)), \textrm{with}\;
    \phi_i = 2\pi\left[\frac{i}{\Phi}\right], z_i = 1 - \frac{2i + 1}{n},
    \label{eq:sp}
\end{equation*}
where $\left[x\right]$ is the fractional part of 
$x: \left[x\right] = x - \lfloor x \rfloor$, $C$ is the conversion of unit vectors from polar to Cartesian coordinates $C(\theta, \phi) = (x,y,z)^T = (\cos(\phi) \sin(\theta), \sin(\phi) \sin(\theta), \cos(\theta))^T$ and $i \in \left\{0, \dots,  n-1\right\}$.
We have two SF~point sets~$\textrm{P}^M_{o}$ and ~$\textrm{P}^N_{d}$, where
$M$ is the number of ray origins $o$ and $N$ is the number of directions $d$.
The set of all rays $\textrm{R}^K$ is determined by the two spherical Fibonacci point sets~$\textrm{P}^M_{o}$ and ~$\textrm{P}^N_{d}$.
The each-to-each connection of $\textrm{P}^M_{o}$ and ~$\textrm{P}^N_{d}$ yields $M \times N$ as the cardinality of $\textrm{R}^K$. 
A ray $r_k \in \textrm{R}^K$ acts as camera ray for the path tracing and is given by 
\begin{equation*}
r_k = (O + Rp^o_i) + t \hat{d}_{i,j}, \textrm{with } d_{i,j} = p^d_j - p^o_i, 
i \in \left\{0, \dots,  M\!-\!1\right\}, j \in \left\{0, \dots,  N\!-\!1\right\}.
\end{equation*}

We use a conventional path tracer comparable to~\cite{pharr2016physically}.
During the tracing, ray origins are uniformly jittered on a disk with area $A = \sfrac{A_{\text{\footnotesize \textbf{S}}}}{M}$ to substantially reduce rendering noise at the cost of additional blurring of the result.
The captured radiance for each $r_k$ is stored in a two dimensional texture.
Each dimension of the texture corresponds to one of the point sets $\textrm{P}^M_{o}$ and ~$\textrm{P}^N_{d}$ and thus, the indices $i$ and $j$ not only identify a point given by the Fibonacci sequence but also the texel coordinates for memory accesses.
Therefore, our reparameterized, discrete form of the plenoptic function is in fact 2D (parameterized by 2 indices of the point sets), denoted by $\hat{L}(i,j)$. 

To synthesize images on the embedded device, we retrieve the precomputed physically-based rendering result from $\hat{L}(i,j)$ during the rasterization process of very simple sphere geometry, which is a simple texturing process~(see Fig.~\ref{fig:image_generation}
%, backface culling enabled
). 
%Alternatively a screen space aligned quad could be used).  
For each rasterization ray that hits the sphere~\textbf{S}, we find two hit points, $h_o$ and $h_d$ for the front and back face, respectively (discarding tangential rays). 
Given a point $h$ on the sphere~\textbf{S}, we use Keinert's inverse mapping~\cite{KeinertACMTG2015} to find the nearest neighbor in an SF point set~P in constant time.
Hence, sampling of $\hat{L}$ queries the nearest neighbor of $h_o$ from $\textrm{P}_o^M$ and $h_d$ from $\textrm{P}_d^N$, denoted by $i$ and $j$, and retrieves the sampled value $\hat{L}(i, i)$ from our plenoptic function. 

\begin{figure}[tb]
\centering
\includegraphics[width=\textwidth]{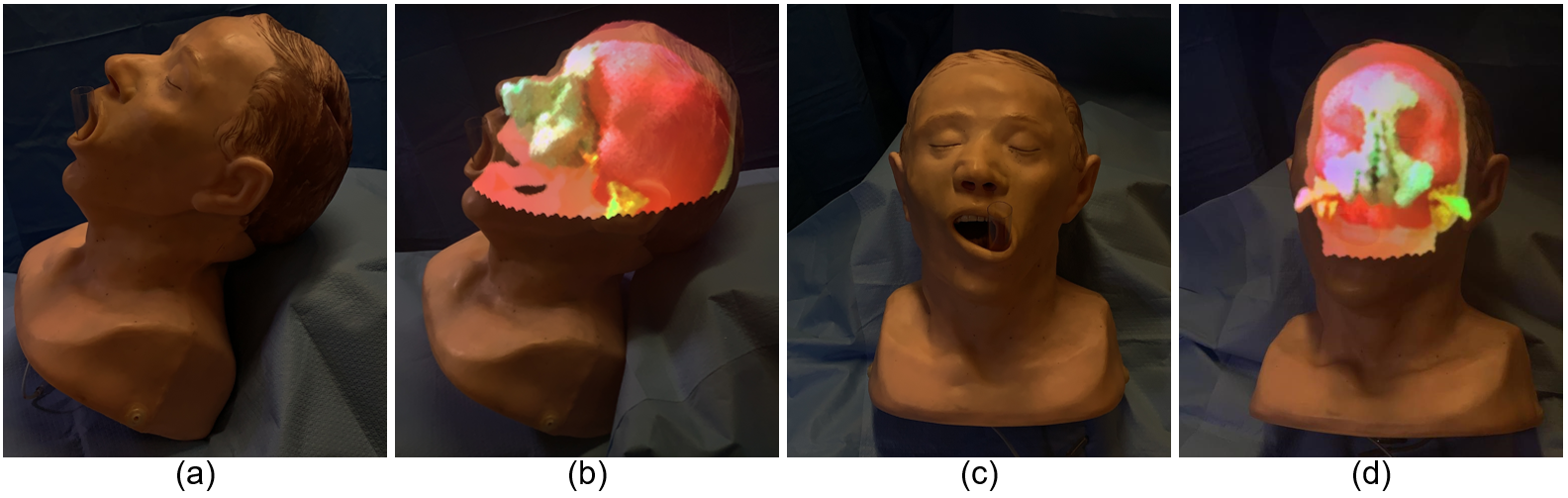}\vspace{-10pt}
    \caption{A human head phantom without overlay (a,c) augmented with representative LumiPath-based renderings without post processing (c,d).
    \label{fig:results-holo}}
\end{figure}

Unfortunately, nearest neighbor sampling yields images that are piece-wise constant and thus, unpleasant in appearance. 
We modify the query to return up to nine neighbors of a point~$h$ instead.
%At the cost of additional memory accesses, this allows for taking into account multiple samples of $\hat{L}$.
We observe that considering five neighbors for each point $h_o$ and $h_d$ leads to sufficient results, for 25 samples of $\hat{L}$ per pixel of the displayed image. 
Neighbors are weighted by their distance to the original hitpoints via a filter kernel of size $R\sqrt[4]{5}\sqrt{\frac{4\pi}{\sqrt{5}N}}$, where $N$ is the size of the SF point set. 
As the inverse mapping has constant time complexity and the number of samples is fixed, our image generation algorithm has constant time complexity for each pixel. 
Fig.~\ref{fig:results-holo} shows representative images obtained with this method. 

\vspace{3mm}
\noindent\textbf{Generative Adversarial Network-based Post Processing} 
Analyzing the higher frequency components of a LumiPath-based image clearly reveals a deterministic pattern of artifacts as shown in Fig.~\ref{fig:results-views}. 
A known method to improve image-based renderings is a view point or parallax correction which takes into account the distance of a hitpoint to the rendered surface~\cite{GortlerSIGGRAPH1996}; however, this is non-trivial in use cases such as volume rendering, where the depth of hitpoints is ill-defined.

We use non-linear filtering in the form of a generative adversarial network (GAN) to improve image quality. %Encouraging progress in this regard has been made in the field of movie restoration and game remastering. 
Our network structure is adapted from \cite{Isola2017}. We use a 3-layer U-Net as generator. The generative loss is the weighted sum of the \textit{Structural Similarity Index}~(SSIM) and L2 loss to encourage smooth structural and color reconstruction. The discriminator is made of 7 blocks of 2D convolution followed by ReLU and dropout. We modify the discriminator's last layer to be average pooling, which emphasizes local patterns, as observed in our artifacts. We use the relativistic GAN, which assigns confidence value to whether a sample is fake or real, with mean-squared loss to speed up convergence~\cite{jolicoeur2018relativistic}.
The renderings of our dataset were generated from alternating viewpoints with uniformly sampled distances and view angles withing reasonable ranges and facing the object of interest. The renderings were randomly distributed into  train~(2220 image pairs), validation~(204 image pairs) and test set~(195 image pairs).
Areas that the LumiPath did not cover were masked out in the reference image during training to prevent the network from hallucinating missing image parts and rather concentrate on local artifact patterns.

\section{Results}
\label{sec:result}

\begin{center}
\begin{figure}[tb]

\begin{minipage}[c]{0.23\linewidth}
\centering
\includegraphics[width=\textwidth]{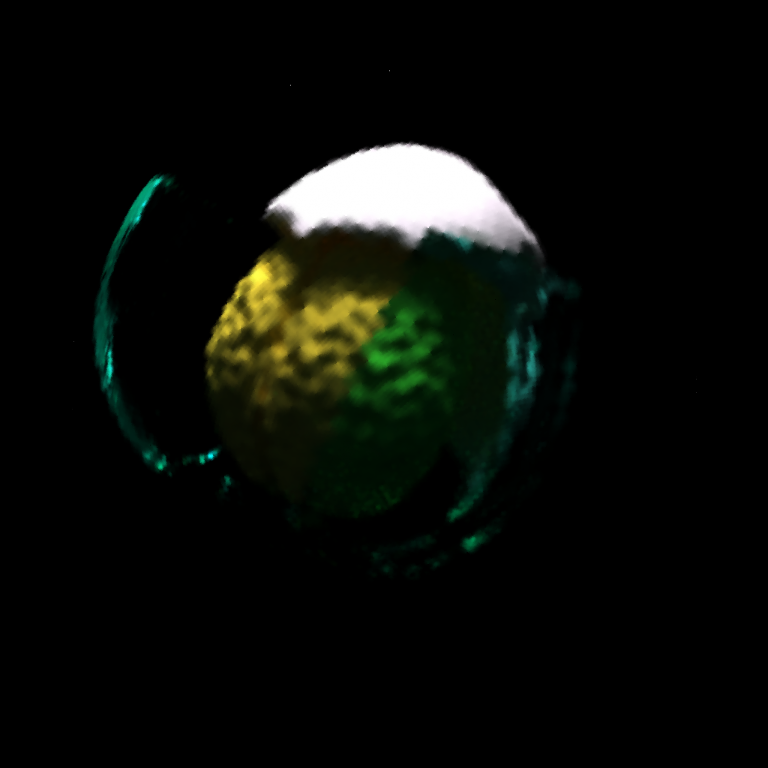}
\end{minipage}
\begin{minipage}[c]{0.23\linewidth}
\centering
\includegraphics[width=\textwidth]{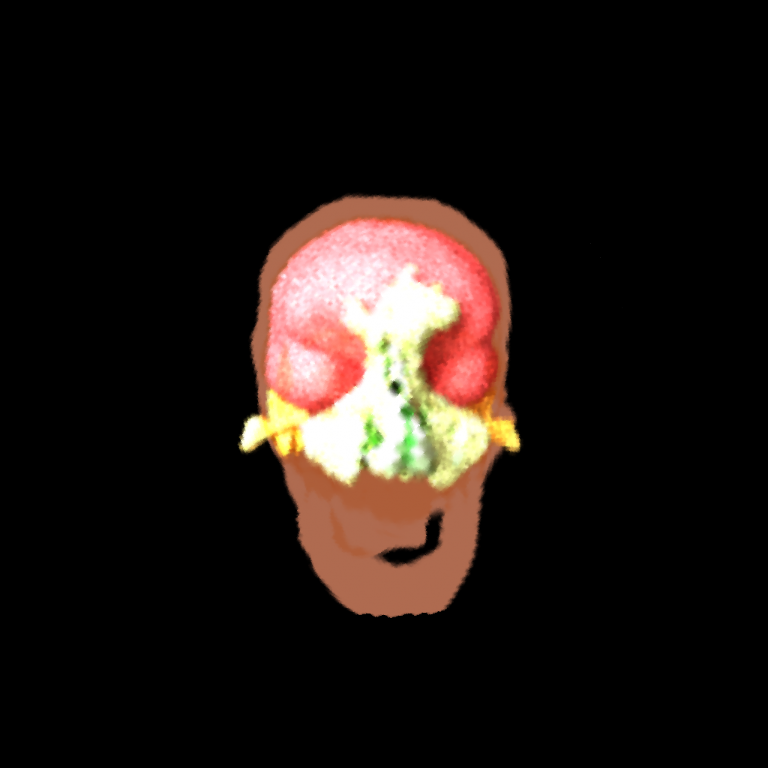}
\end{minipage}
\begin{minipage}[c]{0.23\linewidth}
\centering
\includegraphics[width=\textwidth]{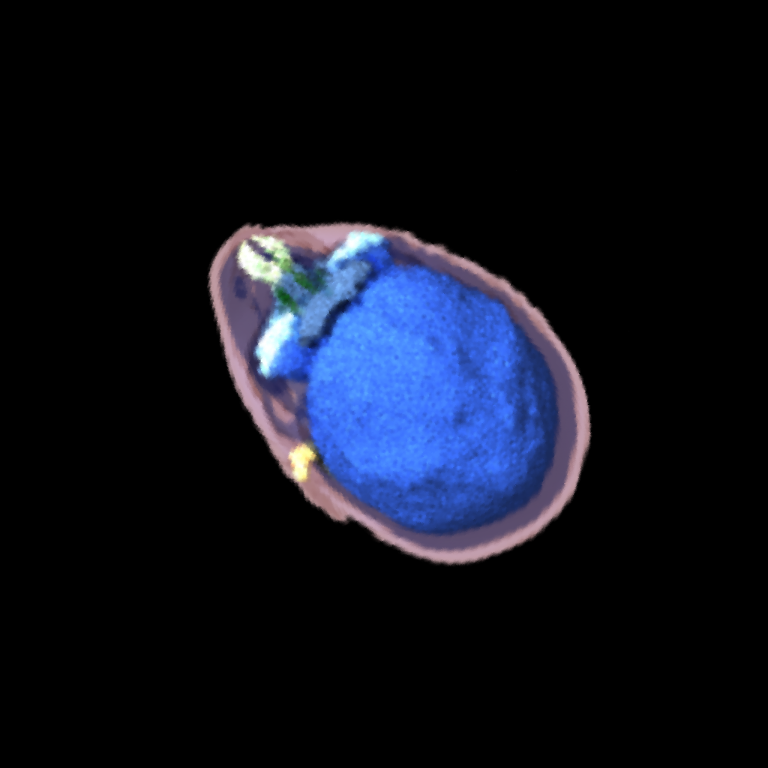}
\end{minipage}
\begin{minipage}[c]{0.23\linewidth}
\centering
\includegraphics[width=\textwidth]{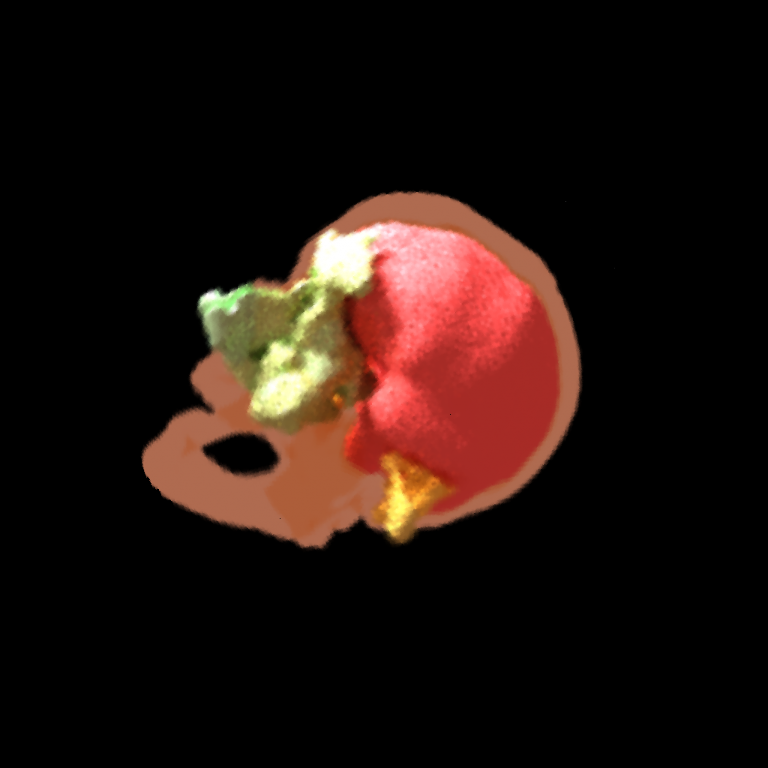}
\end{minipage}

\begin{minipage}[c]{0.23\linewidth}
\centering
\includegraphics[width=\textwidth]{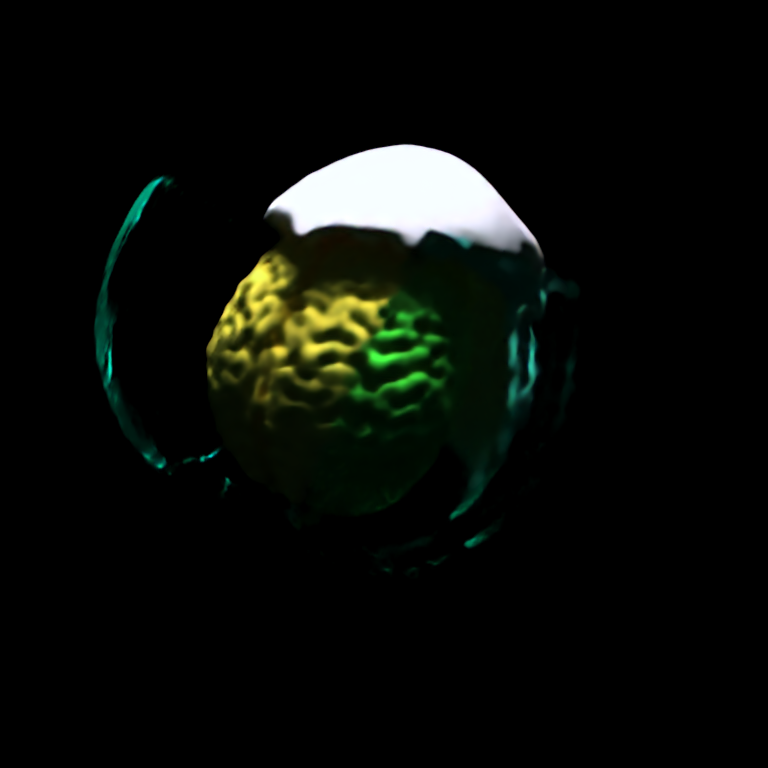}
\end{minipage}
\begin{minipage}[c]{0.23\linewidth}
\centering
\includegraphics[width=\textwidth]{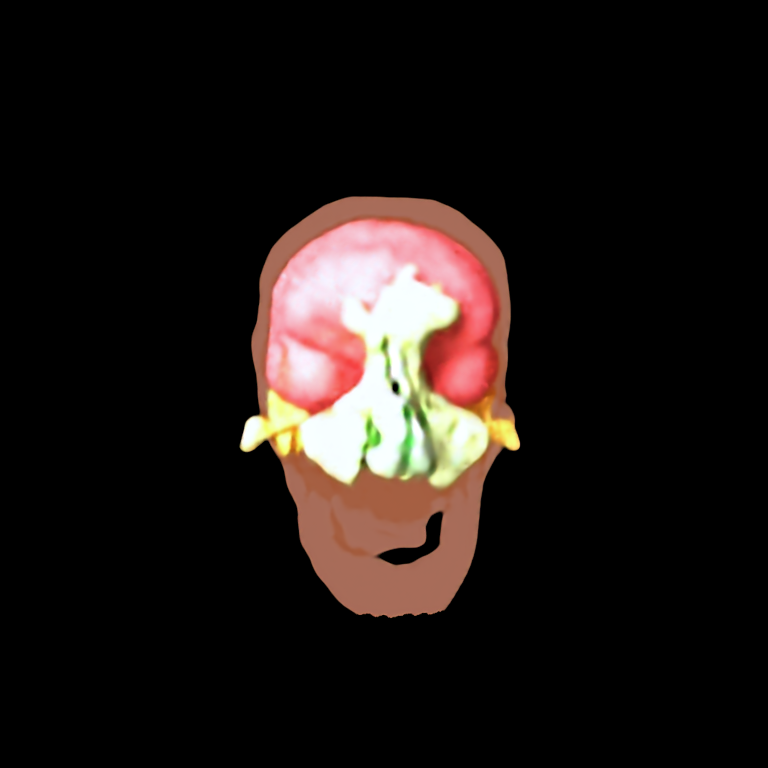}
\end{minipage}
\begin{minipage}[c]{0.23\linewidth}
\centering
\includegraphics[width=\textwidth]{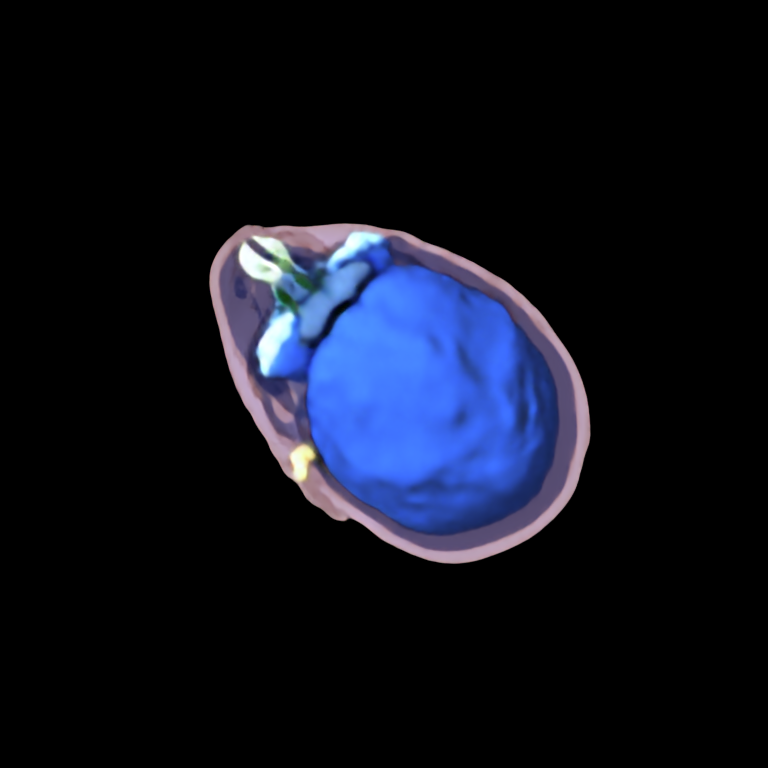}
\end{minipage}
\begin{minipage}[c]{0.23\linewidth}
\centering
\includegraphics[width=\textwidth]{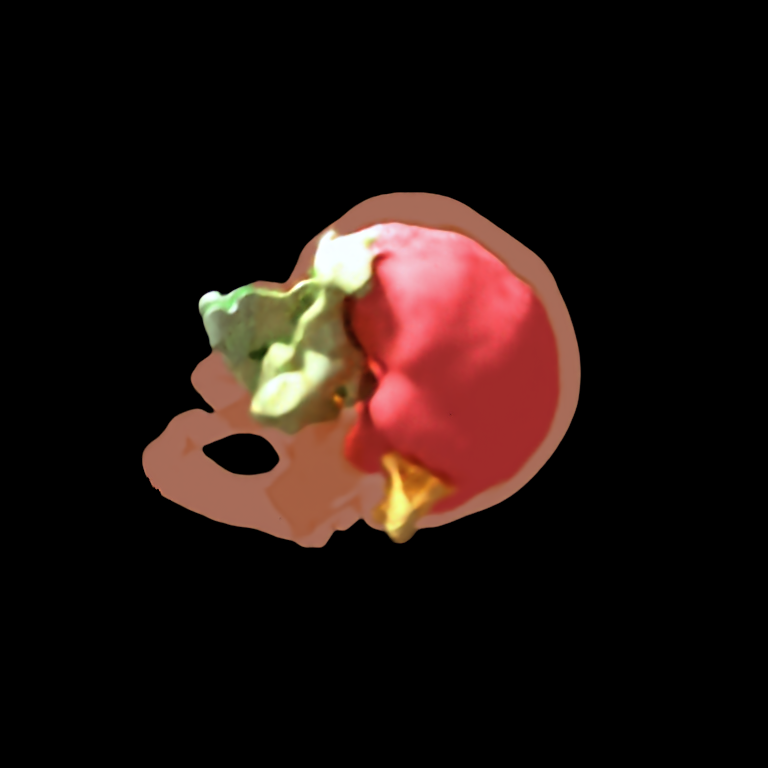}
\end{minipage}

\begin{minipage}[c]{0.23\linewidth}
\centering
\includegraphics[width=\textwidth]{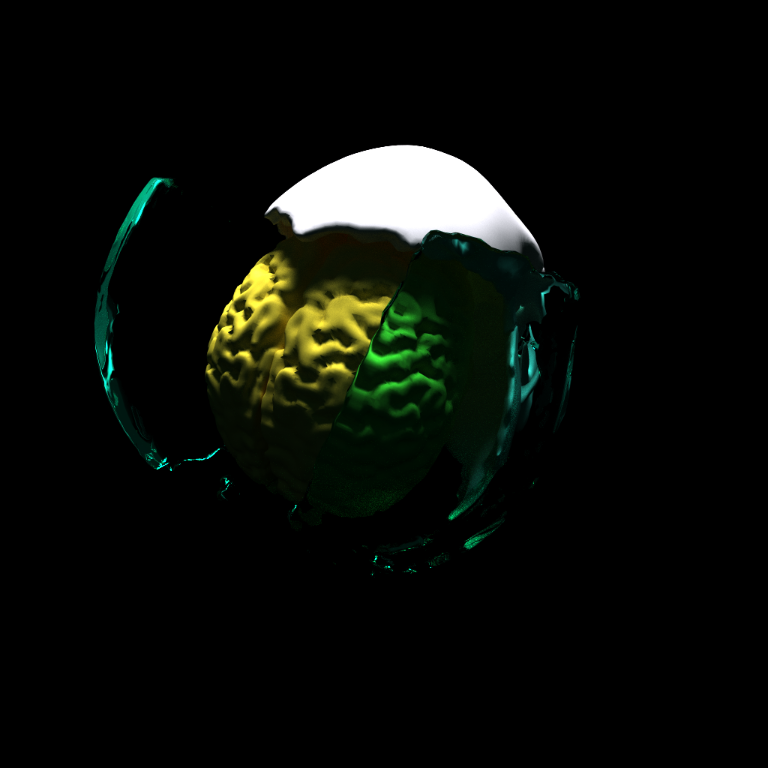}
\end{minipage}
\begin{minipage}[c]{0.23\linewidth}
\centering
\includegraphics[width=\textwidth]{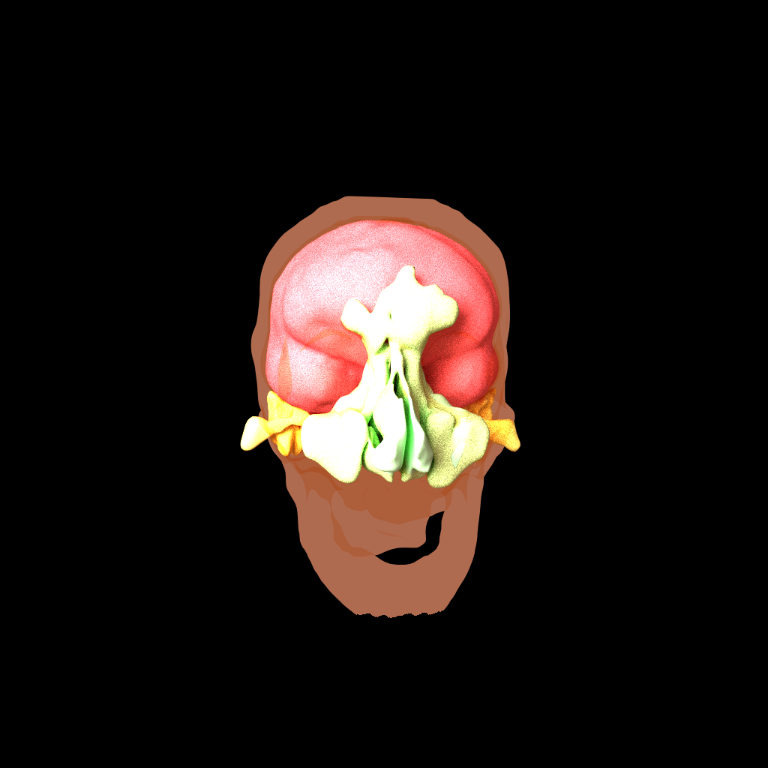}
\end{minipage}
\begin{minipage}[c]{0.23\linewidth}
\centering
\includegraphics[width=\textwidth]{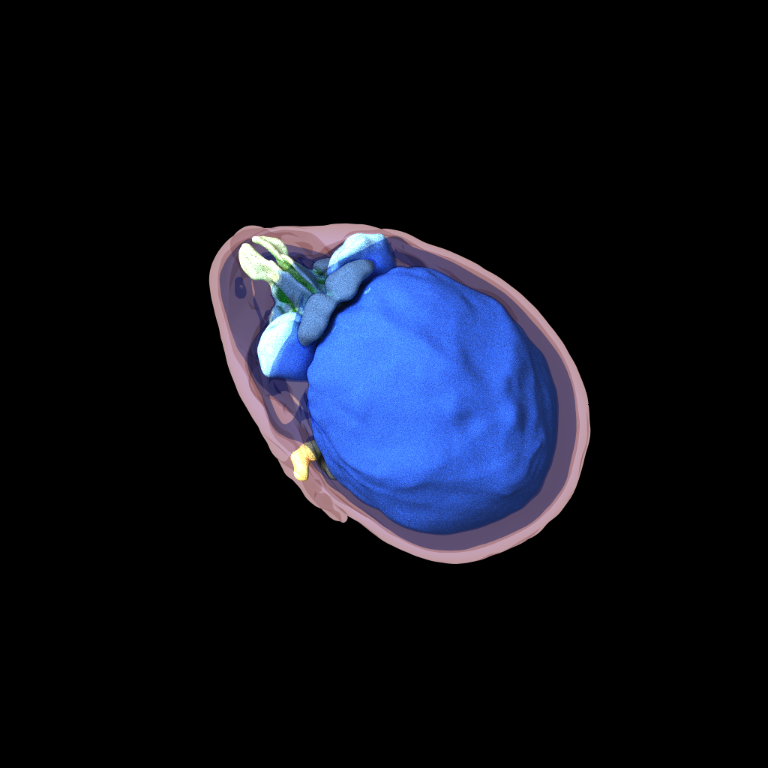}
\end{minipage}
\begin{minipage}[c]{0.23\linewidth}
\centering
\includegraphics[width=\textwidth]{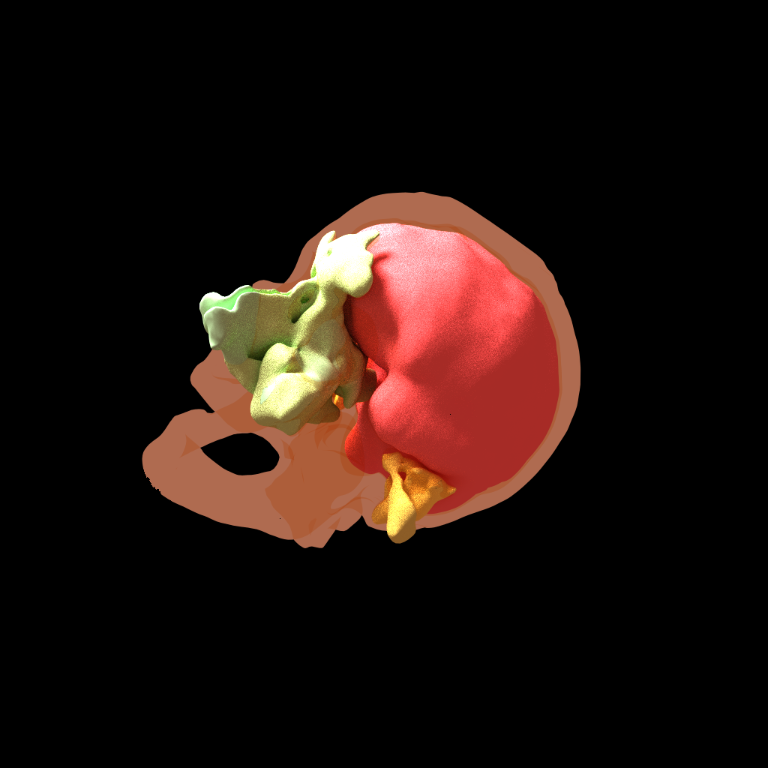}
\end{minipage}

    \caption{Sample rendering results for four views of our test objects (segmented surfaces from CT data and skull model~\cite{skullmodel}). Top: Lumipath-based without post processing~(pp). Middle: Lumipath-based with pp. Bottom: Conventionally path traced. 
    }
    \label{fig:results-views}
\end{figure}
\end{center}

Fig.~\ref{fig:results-views} shows representative images that were synthesized using a conventional path tracer and our \emph{LumiPath} plenoptic function with and without learning-based post-processing. 
We evaluated our rendering results with respect to the conventionally path traced image (1024 samples per pixel) using quantitative image quality metrics, namely the SSIM and the \textit{Complex Wavelet SSIM}~\cite{sampat2009complex}, which are commonly used for image quality assessment. 
In contrast to the SSIM, the CWSSIM performs a complex wavelet transform of the image to a steerable pyramid (with 8 levels in total) prior to the analysis of contrast, structure and luminance~\cite{sampat2009complex}. 
Therefore, the CWSSIM is especially interesting as image-quality trade-offs for our LumiPath-based renderings are most prominent in higher frequency domains, e.\,g. along edges and specular highlights~(see Fig.~\ref{fig:results-views}). Considering ten representative views, we obtain average SSIM values of 0.972/0.975 and a CWSSIM of 0.997/0.998 with/without post-processing post-processing. 

The cardinality of the point sets $\textrm{P}^M_{o}$ and ~$\textrm{P}^N_{d}$ were set to $M = 12288$ and $N = 23576$, with points of $\textrm{P}^M_{o}$ limited to the upper hemisphere.
Consequently, the precomputed texture had a size of 576\,MB~($0.5\times12288\times23576\times4$~Byte).
We evaluated the performance of our (na\"ive) prototype on the Microsoft Hololens~v1. We measured framerates between $\approx 5.2$ to $9.1$\,fps ($\approx 14$ to $15$\,fps in the emulator) for both eyes in total and views similar to the ones shown in Fig.~\ref{fig:results-holo}.
In comparison, rendering one frame of our ground truth as shown in Fig.~\ref{fig:results-views} took about 3~minutes on a \textit{Nvidia GTX 980M}.
Our path tracing framework
%to generate the ground truth as well as the LumiPath texture 
is based on the \textit{Nvidia Optix Engine}.

\section{Discussion \& Conclusion}
\label{sec:discussion}
We present first steps towards a physically-based rendering pipeline on embedded devices that show promising results. The proposed method achieves $\approx7.5$\,fps for both views on a Hololens~v1. More work to optimize the code and enable GPU use will further improve performance. 

As our prototype is currently designed in two disjoint parts, rendering is limited to static objects. In case of changing conditions, the light field has to be recalculated. Further, our current prototype visualizes surfaces rather than volumes. We will investigate how our approach translates to volume rendering applications that, ultimately, we consider our method most useful for.

%Further research to incorporate our machine learning-based post-rendering correction into our existing framework has to be done, so that its actual performance can be evaluated on embedded devices.
While both the rendering and network run on the HoloLens, limited memory restricts them to run sequentially, with network execution not currently real-time capable. This can be mitigated by further code optimization, but as embedded devices become more powerful, their increased memory bandwidth and dedicated tensor processing units will enable more concurrency.
We show that using a generative network to perform non-linear filtering, we can remove artifacts from our interpolation method. Our renderings were based on a lightfield that was below $600$\,MB, which seems appropriate for today's embedded devices. Additionally, during the quantitative comparison to the pathtraced ground truth, we observe that our network implicitly denoises our rendering, which further enhances the perceived quality. 

Future work in post-processing may explore network architectures that directly sample from our plenoptic function to synthesize the desired image and can be trained end-to-end. Such approach would be appealing since the origin and direction
%$o$ and $d$ 
can be taken into account during the inference process. Doing so is not possible with the post-rendering correction described here that operates on already interpolated color values. In addition, a network that is aware of our light field structure might be helpful to further reduce the number of texture accesses, which we observed as one of the biggest bottlenecks in our current prototype (texture reads made up $\approx$33.3\,\% of the frametime). 

While we currently use machine learning-based post-rendering corrections, other approaches can also be incorporated into our pipeline to improve image quality. A promising approach would be the investigation of non-uniform sampling patterns of the Fibonacci Spheres, e.\,g. based on specific object properties.
%such as curvature. 
However, adaptive sampling of Fibonacci Spheres is complex and requires sophisticated handling of boundaries. 

In summary, we understand our results as promising, yet preliminary evidence that our LumiPath algorithm can achieve reasonable real-time physically-based rendering results on untethered compute-limited devices such as OST HMDs. Finally, these developments may prove useful for light field displays which bring dynamic focal lengths to OST HMDs and solve the vergence-accommoda\-tion conflict~\cite{kramida2015resolving}.

\vspace{3mm}
\noindent \textbf{Acknowledgements } \ifanonymized No acknowledgements given at this time. \else The Titan V used for this research was donated by the Nvidia Corporation. The authors would like to thank Benjamin Keinert for helping to understand the inverse Fibonacci mapping and Arian Mehrfard for his help in acquiring screenshots. \fi

%\bibliographystyle{splncs04}
%\bibliography{bibliography}
%

\end{document}